\documentclass[conference]{IEEEtran}
\usepackage[utf8]{inputenc}
\usepackage{microtype}
\usepackage{amsmath,amssymb,bm,mathtools,stackrel}
\usepackage{graphicx}
\usepackage{blindtext}
\usepackage{cite}
\usepackage{booktabs}

\usepackage{tikz}
\usetikzlibrary{calc,fit}

\usepackage{pgfplots}

\begin{document}
\title{SPN-CNN: Boosting Sensor-Based Source Camera Attribution With Deep Learning}

\author{%
\IEEEauthorblockN{Matthias Kirchner and Cameron Johnson}
\IEEEauthorblockA{Kitware, Inc.\\ 
Email:  \{matthias.kirchner,cameron.johnson\}@kitware.com}
}


\maketitle

\begin{figure}[b]
\vspace{-0.3cm}
\parbox{\hsize}{\textcolor{black}{\em
WIFS`2019, December, 9-12, 2019, Delft, Netherlands.\\
Approved for public release: distribution unlimited.\\\copyright\,2019 IEEE. }
}\end{figure}

\begin{abstract}
We explore means to advance source camera identification based on sensor noise in a data-driven framework. Our focus is on improving the sensor pattern noise (SPN) extraction from a single image at test time.  Where existing works suppress nuisance content with denoising filters that are largely agnostic to the specific SPN signal of interest, we demonstrate that a~deep learning approach can yield a more suitable extractor that leads to improved source attribution. A series of extensive experiments on various public datasets confirms the feasibility of our approach and its applicability to image manipulation localization and video source attribution. A critical discussion of potential pitfalls completes the text.
\end{abstract}

\IEEEpeerreviewmaketitle

\section{Introduction}
\label{sec:intro}

Sensor noise fingerprints have been recognized and utilized as a cornerstone of media forensics \cite{Bohme:2016aa} ever since Luk\'a\v{s} et al. first observed almost 15 years ago in their seminal work \cite{Lukas:2005ab} that digital images can be traced back to their sensor based on unique noise characteristics. Minute manufacturing imperfections are believed to make every sensor physically unique, leading to the presence of a weak yet deterministic sensor pattern noise (SPN) in each and every image signal captured by the same sensor \cite{Fridrich:2013dq}. This fingerprint, commonly referred to as \emph{photo-response non-uniformity} (PRNU), can be estimated from images captured by a specific camera for the purpose of source camera identification, in which a noise signal extracted from a probe image of unknown provenance is compared against pre-computed fingerprint estimates from a set of candidate~cameras. 

While extensive testing has already demonstrated the feasibility of highly reliable PRNU-based consumer camera identification at scale~\cite{Goljan:2009aa}, recent research has been largely driven by enabling source attribution under ever more challenging conditions. Modern cameras, particularly those installed in smartphones, go to great lengths to produce visually appealing imagery, and techniques such as lens distortion correction, electronic image stabilization, or high dynamic range imaging have all been found to impede camera identification if not accounted for through spatial resynchronization \cite{Goljan:2012aa,Mandelli:2019aa,Hosseini:2019aa}. Robustness to strong image and video compression is another major concern \cite{Houten:2009ab,Chuang:2011aa,Goljan:2016aa} that has been gaining more and more practical relevance due the widespread sharing of visual media through online social networks \cite{Amerini:2017aa}. Finally, the ability to reliably establish camera identification also from very small image patches is crucial for image manipulation localization based on sensor noise \cite{Chen:2008aa,Chierchia:2014ab,Korus:2017aa}.

The pertinent literature concludes that all these scenarios generally benefit from high-quality fingerprint estimates, and that images should undergo content suppression prior to analysis \cite{Fridrich:2013dq}. To date, most works still resort to the Wavelet-based denoiser as adopted by Luk\'a\v{s} et~al.~\cite{Lukas:2006aa} for that purpose. The maximum likelihood fingerprint estimator derived from a simplified multiplicative signal model \cite{Chen:2008aa} gives near-optimal results when fed with noise residuals from a set of full-resolution images of a homogeneously lit scene. The conditions at test time are less ideal, however. Probe images are generally of varying content and an aggregation over multiple images is often not possible. It is accepted that noise residuals obtained from off-the-shelf denoisers are imperfect by nature and that they are contaminated non-trivially by remnants of image content. Salient textures or quantization noise exacerbate the issue. A number of studies have found that alternative denoising algorithms can lead to moderate improvements \cite{Amerini:2009aa,Cortiana:2011aa,Chierchia:2014ab}, yet there remains a considerable gap in the ability to reliably establish the presence of the camera fingerprint in (portions of) a probe image under more challenging conditions. 

While the overall performance of camera identification is clearly governed by fundamental bounds imposed by, for instance, the available image resolution and the strength of compression, this paper sets out to demonstrate that there is still ample room for advances over prior art. Critically, the present work deviates from the common procedure to employ at test time the very same noise extractor that was also used for fingerprint estimation, which we explain with a subtle shift in perspective in the restatement of practical camera identification: we accept that the goal ultimately is to match a noise signal extracted from the probe image against a pre-computed fingerprint estimate, and that the quality of the match is assessed in terms of the similarity of the two signals. Adopting the DnCNN work by Zhang et al.~\cite{Zhang:2017aa}, 
we let a convolutional neural network (CNN) learn how to extract a noise signal from a probe image that resembles the pre-computed sensor pattern noise (SPN) fingerprint from the camera of interest as closely as possible. By looking at fingerprint extraction through the lens of an optimization procedure, the resulting network, which we call SPN-CNN, can be expected to better adapt to the problem at hand than existing fingerprint-agnostic denoising algorithms. It is worth pointing out here that the proposed technique differs from the creative advancement of the DnCNN idea in Cozzolino and Verdoliva's   NoisePrint approach \cite{Cozzolino:2018aa} both in its objective and its design in that the computed noise signal is expected to emphasize device characteristics instead of camera model characteristics through a training regime that explicitly utilizes a known camera-specific target signal (the camera fingerprint estimate) instead of a more general notion of pairwise patch similarity. Another related work is the image anonymization approach by Bonettini et al. \cite{Bonettini:2018aa}, who adapted a DnCNN-like design to \emph{remove} the camera fingerprint from an image. We refer to Section~\ref{sec:fpcnn} for a more detailed exposition of our approach, which follows after a brief overview of camera sensor noise forensics in Section~\ref{sec:sensor-forensics}. Section~\ref{sec:experiments} presents experimental results, including applications to image manipulation localization and video source identification. Section~\ref{sec:conclusion} concludes the text.


\section{Camera Sensor Noise Forensics}
\label{sec:sensor-forensics}

State-of-the-art sensor noise forensics assumes a simplified imaging model of the form
\begin{equation}
\bm{x} = \bm{x}^{(o)} (\bm{1}+\bm{k}) + \bm{\theta}\, ,
\end{equation}
in which the multiplicative PRNU factor $\bm{k}$ modulates the noise-free image $\bm{x}^{(o)}$, while $\bm{\theta}$ comprises a variety of additive noise components \cite{Fridrich:2013dq}. 
Substantial empirical evidence suggests that the PRNU is a unique and robust camera fingerprint \cite{Goljan:2009aa} that can be estimated from a set of $N$ images taken with the specific camera of interest. The standard procedure relies on a denoising filter $F(\cdot)$ to obtain the noise residual $\bm{w}_n =\bm{x}_n- F(\bm{x}_n)$ from the $n$-th image $\bm{x}_n$, $1\leq n\leq N$. A modeling assumption
\begin{equation}
\bm{w}_n = \bm{k}\bm{x}_n + \bm{\eta}_n
\end{equation}
with i.\,i.\,d.~Gaussian noise $\bm{\eta}_n$ then leads to a maximum likelihood estimate $\hat{\bm{k}}$ of the PRNU factor of the form \cite{Chen:2008aa}
\begin{equation}
\hat{\bm{k}} = \left(\,\sum_{n=1}^N \bm{w}_n \bm{x}_n\right)\cdot \left(\, \sum_{n=1}^N \bm{x}_n^2 \right)^{-1}\,
\label{eq:ml-estimate} .
\end{equation}
Practical applications warrant a post-processing step to clean the fingerprint estimate from non-unique artifacts \cite{Fridrich:2013dq,Gloe:2012pt}.

For a given probe image $\bm{y}$ of unknown provenance, camera identification is formulated as a hypothesis testing problem:
\begin{align*}
H_0:{} & \bm{w} = \bm{y}- F(\bm{y}) \text{ does not contain the fingerprint } \bm{k}\\
H_1:{} & \bm{w} \text{ does contain the fingerprint } \bm{k}\, ;
\end{align*} 
i.\,e., the probe is attributed to the tested camera if $H_1$ holds. In practice, the test can be decided by evaluating the similarity between the residual $\bm{w}$ and the fingerprint estimate $\hat{\bm{k}}$ for a suitably set threshold $\tau$,
\begin{equation}
\rho = \mathrm{sim}(\bm{w}, \hat{\bm{k}}) \stackrel[H_0]{H_1}{\gtrless} \tau\, .
\label{eq:objective}
\end{equation}
It is assumed that the two signals are geometrically aligned except for possible translational displacements, which can be accounted for conveniently with normalized cross-correlation or peak-to-correlation energy (PCE) as similarity measures \cite{Fridrich:2013dq}.

As inserting content from elsewhere into an image or other forms of image manipulation will remove or impair the camera fingerprint in the affected regions, many local image alterations can be detected by testing for the presence of the expected fingerprint in a sliding window mode \cite{Chen:2008aa}. The localization of small manipulated regions warrants sufficiently small analysis windows, which generally impacts the ability to reliably establish whether or not the expected fingerprint is present negatively.  The literature often recommends a window size of about $64\times 64$ pixels as a reasonable trade-off between resolution and accuracy \cite{Chierchia:2014ab,Chakraborty:2017aa,Korus:2017aa}. A core problem is that the measured local similarity scores under $H_1$ depend greatly on local image characteristics. Chen et al.~\cite{Chen:2008aa} have proposed a correlation predictor $\hat{\rho}_i(\bm{x})$ as a remedy, which utilizes a set of simple intensity and texture features to predict how strongly the $i$-th local patch in a probe image \emph{would} correlate with the purported camera fingerprint under $H_1$. The decision whether to declare a manipulation (i.\,e., the absence of the expected fingerprint) can then be made based on the deviation from the expected correlation.

As for video data, it can be advantageous to estimate reference sensor fingerprints from full-resolution still images when available \cite{Iuliani:2019aa}. At test time, it is usually recommended to aggregate noise residuals from multiple probe frames into a probe video fingerprint to cope with strong compression artifacts \cite{Chen:2007ab}. Special care has to be taken to account for geometrical desynchronization due to video stabilization, as pointed out in many recent reports \cite{Taspinar:2016aa,Iuliani:2019aa,Mandelli:2019aa}.

\section{SPN-CNN Estimator}
\label{sec:fpcnn}

While camera source attribution from sensor pattern noise arguably works extremely well iff pitfalls due to geometric misalignment between the fingerprint and the probe are avoided, there remains a huge discrepancy between a fingerprint estimate aggregated from several images,~$\hat{\bm{k}}$, and an estimate from a single probe image (via the residual~$\bm{w}$). A simple explanation is that denoising algorithms are not designed with this specific application in mind, and more generally that content suppression is an ill-posed problem in the absence of  viable image models (and possibly even of the noise characteristics itself~\cite{Masciopinto:2018aa}). Thanks to the central limit theorem, aggregation over multiple noise residuals mitigates these effects to some extent, but noise residuals from a single image will always suffer from significant distortion.

The recent success of data-driven methods suggests a way forward, however, especially also considering that Equation~\eqref{eq:objective} poses a clear objective function to be targeted. By understanding CNNs as flexible non-linear optimization tool, it seems reasonable to expect that a network should be able to learn how to extract a better approximation of $\bm{k}$ from a given probe than a conventional ``blind'' denoiser. Conceptually, the problem at hand aligns with the scenario considered by Zhang et al.~\cite{Zhang:2017aa}, who trained a CNN to learn how to extract well-characterized noise signals from a given image. In our case, we train a network to extract a noise pattern $\tilde{\bm{k}}$ to minimize $\|\hat{\bm{k}} - \tilde{\bm{k}}\|_2^2$.
Our premise here is that the maximum likelihood fingerprint estimator (MLE) gives the best approximation of the actual PRNU signal that we have under the given imaging model assumptions, so we consider it a viable proxy target in lieu of the unknown ground truth. Once trained, the network replaces the denoiser $F(\cdot)$ at test time, i.\,e., $\bm{w} = \tilde{\bm{k}}$ in Equation~\eqref{eq:objective}. Notably, this breaks with the tradition of employing the very same denoiser for both fingerprint estimation and detection, although further iterations in which the trained network informs a better fingerprint aggregation seem conceivable. 

\begin{figure}
\includegraphics[width=\linewidth]{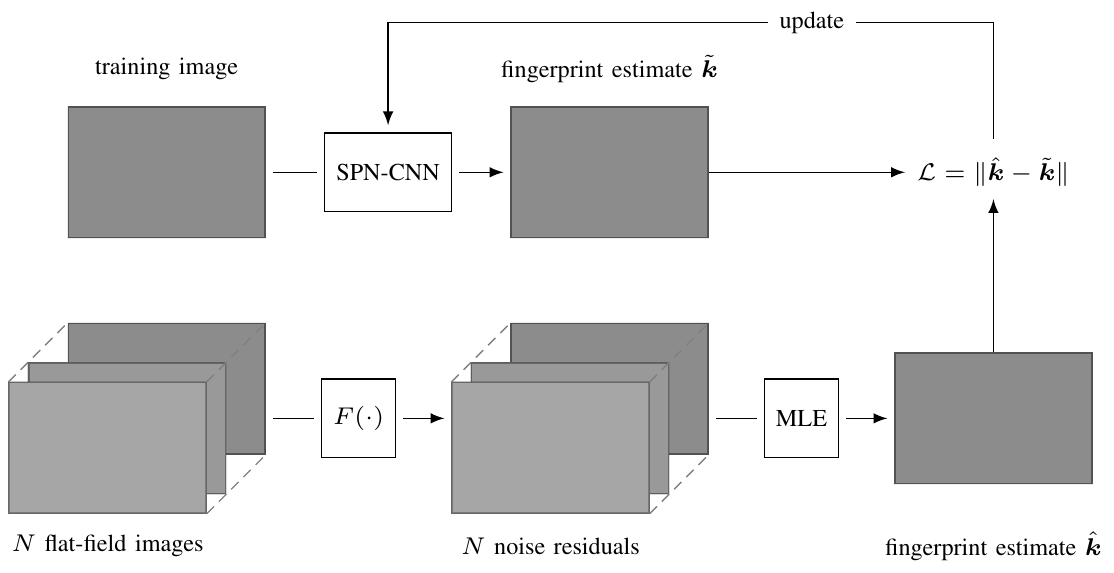}
\caption{Proposed SPN-CNN training for learning to extract a camera fingerprint $\tilde{\bm{k}}$ from a given probe image.}
\label{fig:spn-cnn}
\end{figure}

Figure~\ref{fig:spn-cnn} outlines the overall training setup, which operates, in its current instantiation, on single-channel grayscale patches of size $40\times 40$ pixels. To facilitate the learning, the output from each training patch is paired with its corresponding portion of the fingerprint estimate~$\hat{\bm{k}}$. The SPN-CNN itself is modeled after Zhang et~al.'s DnCNN  \cite{Zhang:2017aa} and comprises 17~layers. Each layer implements 64 convolutional filters with a small $3\times 3$ spatial support, except for the last layer with only a single-channel output. Batch normalization is implemented in between the convolutional layers, and ReLu activation follows after each but the last layer. No pooling is used at any point. In line with reports in the literature, any modern off-the-shelf denoiser $F(\cdot)$  can be expected to do reasonably well for the purpose of estimating a camera fingerprint from flat-field images. We decided to stick with the standard Wavelet formulation \cite{Lukas:2006aa} in this work for its computational edge over alternative approaches. The trained network can be fed with inputs of any size within the memory constraints of the GPU. In practice, we divide images into tiles of size $400\times 400$ with an overlap of 20 pixels to accommodate for possible boundary~effects. 

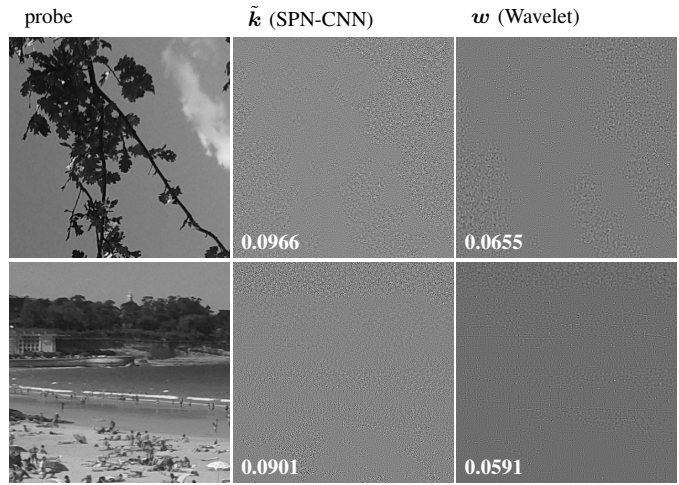
\begin{figure}
\parbox{.33\linewidth}{\footnotesize ~~probe}\hfill%
\parbox{.33\linewidth}{\footnotesize ~~$\tilde{\bm{k}}$ (SPN-CNN)}\hfill%
\parbox{.33\linewidth}{\footnotesize ~~$\bm{w}$ (Wavelet)}\\[3pt]
\begin{tikzpicture}[every node/.style={inner sep=0pt},font=\footnotesize]
\node (X) {\includegraphics[width=.33\linewidth]{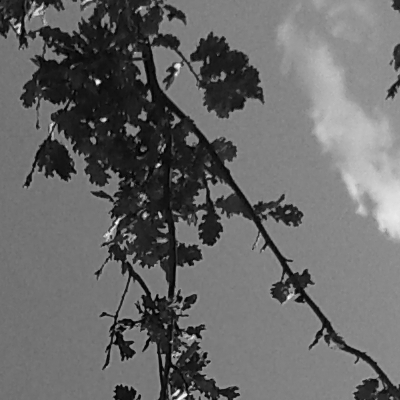}};
\node[right=1pt] (Y) at (X.east) 
	{\includegraphics[width=.33\linewidth]{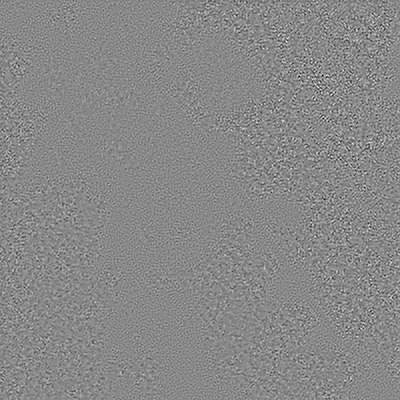}};
\node[right=1pt] (Z) at (Y.east) 
	{\includegraphics[width=.33\linewidth]{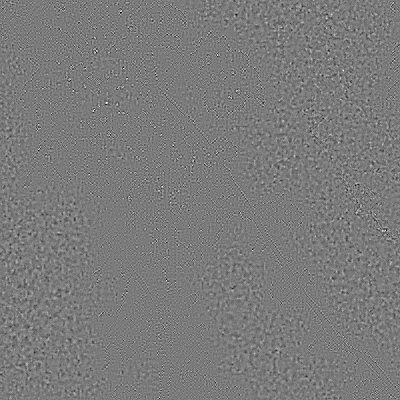}};
\begin{scope}[every node/.style={white,inner sep=3pt,anchor=south west}]
\node at (Y.south west) {\bfseries 0.0966};
\node at (Z.south west) {\bfseries 0.0655};
\end{scope}
\end{tikzpicture}\\[2pt]
\begin{tikzpicture}[every node/.style={inner sep=0pt},font=\footnotesize]
\node (X) {\includegraphics[width=.33\linewidth]{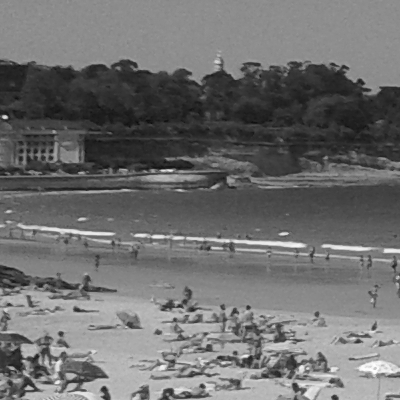}};
\node[right=1pt] (Y) at (X.east) 
	{\includegraphics[width=.33\linewidth]{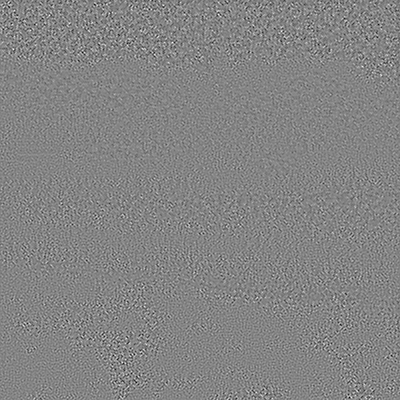}};
\node[right=1pt] (Z) at (Y.east) 
	{\includegraphics[width=.33\linewidth]{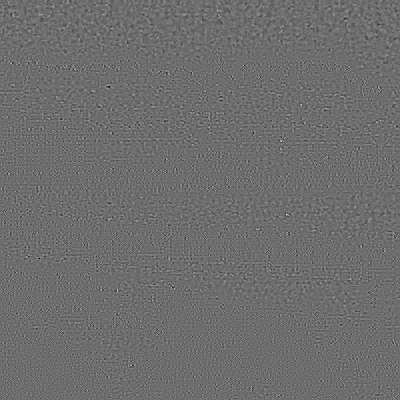}};
\begin{scope}[every node/.style={white,inner sep=3pt,anchor=south west}]
\node at (Y.south west) {\bfseries 0.0901};
\node at (Z.south west) {\bfseries 0.0591};
\end{scope}
\end{tikzpicture}
\caption{Two $400\times 400$ images and the noise signals obtained with the SPN-CNN and Wavelet estimators. Each noise pattern has been scaled independently to cover the full intensity range. Numbers in the lower left corner indicate the correlation with the corresponding Wavelet-based camera fingerprint.}
\label{fig:noise-examples}
\end{figure}

Figure~\ref{fig:noise-examples} depicts exemplary results from two images from the VISION database \cite{Shullani:2017aa} and compares the obtained noise signals, $\tilde{\bm{k}}$, to the corresponding residuals from the Wavelet denoiser, $\bm{w}$. Although the CNN-based estimator does arguably not succeed in suppressing the image content completely, the results suggest qualitatively that it does better than the Wavelet denoiser. A notable increase in correlation with the corresponding portion of the camera fingerprint estimate $\hat{\bm{k}}$ supports this impression.

\section{Experiments}
\label{sec:experiments}

We work with various datasets for an experimental validation. Our \emph{baseline camera identification} results in Section~\ref{sec:baseline} cover camera fingerprints from ten devices from the VISION Database \cite{Shullani:2017aa}, and from six cameras from the Dresden Image Database \cite{Gloe:2010ad}, respectively. The former dataset comprises camera-native JPG images from a variety of mobile devices, whereas we chose the Nikon cameras that provide uncompressed imagery from the latter (converted from raw format with Adobe Lightroom). For each camera, we reserve 100 randomly chosen images for training and divide the remaining images into non-overlapping crops of size $400\times 400$ pixels, resulting in a total of 53,677 (VISION) plus 41,568 (Dresden) cropped test images under $H_1$. Under $H_0$, we supply for each non-overlapping $400\times 400$ portion of the 16 camera fingerprints about 100 randomly chosen image crops from a different device, totaling in 48,362 (VISION) plus 23,004 (Dresden) samples. The \emph{manipulation localization} results in Section \ref{sec:localization} cover the 55 manipulated images from the Nikon D7000 camera in the Realistic Tampering Dataset~\cite{Korus:2017aa}. These uncompressed images originate from the RAISE database \cite{Dang-Nguyen:2015aa}, from which we source an additional 100 images for training. To gain some insight into the applicability to \emph{video source attribution} in Section~\ref{sec:video}, we work with 18 camera-native indoor and outdoor videos from a Samsung Galaxy S3 Mini (D01 in the VISION database), where we reserve five videos for training. All databases provide flat-field images for the initial MLE camera fingerprint estimation with the Wavelet denoiser.

\subsection{Training}
\label{sec:baseline}

We train a separate network for each camera fingerprint of interest in batches of size 128 for 100 epochs. Each epoch sees a selection of non-overlapping $40\times 40$ patches from all training images. The patch selection is greedy and successively samples up to 1,000 random patches per image while rejecting those that overlap with previous selections from within the same image. The final number of selected patches thus depends on the image size, but it typically averages to over 50,000 per epoch. Saturated patches are excluded from training. We use the Adam optimizer with MSE loss for training with learning rate $10^{-3}$ and weight decay of 0.2 after every 30 epochs.

\subsection{Baseline Results}
\label{sec:baseline}

\begin{table}
\caption{Median correlation under $H_1$ ($400\times 400$ patches, VISION).}%
\label{tab:vision}%
\vspace*{-4pt}\resizebox{\linewidth}{!}{\scriptsize%
\begin{tabular}{@{}c@{\ \ \ }*{11}{c@{\ \ }}c@{}}
\toprule
& \multicolumn{10}{c}{SPN-CNN trained for camera} \\
\cmidrule(r){2-11}
\tikz[overlay]\node[rotate=90,xshift=.2cm,yshift=2pt]{camera};
\tikz[overlay]\node[rotate=0,xshift=.5cm,yshift=.55cm]{extractor};
\tikz[overlay]\draw[] (-.2,0.65) -- (.3,-0.1);
    &  D01  & D02   & D03   & D06   & D10   & D15   & D26   & D27   & D29   & D34   & Wav   & \cite{Zhang:2017aa} \\
\cmidrule(r){2-11}
\cmidrule(r){12-13}
D01 & \bf 0.075 & 0.058 & 0.046 & 0.055 & 0.062 & 0.053 & 0.072 & 0.058 & 0.061 & 0.063 & 0.055 & 0.054 \\
D02 & 0.070 & \bf 0.082 & 0.051 & 0.065 & 0.072 & 0.073 & 0.072 & 0.071 & 0.072 & 0.073 & 0.064 & 0.060 \\
D03 & 0.020 & 0.020 & \bf 0.045 & 0.027 & 0.028 & 0.032 & 0.019 & 0.030 & 0.029 & 0.027 & 0.024 & 0.019 \\
D06 & 0.058 & 0.054 & 0.054 & \bf 0.092 & 0.061 & 0.068 & 0.058 & 0.064 & 0.062 & 0.079 & 0.051 & 0.051 \\ 
D10 & 0.065 & 0.057 & 0.044 & 0.052 & \bf 0.082 & 0.051 & 0.059 & 0.056 & 0.061 & 0.059 & 0.056 & 0.051 \\ 
D15 & 0.073 & 0.070 & 0.074 & 0.087 & 0.076 & \bf 0.116 & 0.071 & 0.081 & 0.086 & 0.081 & 0.065 & 0.060 \\
D26 & 0.062 & 0.045 & 0.038 & 0.048 & 0.053 & 0.046 & \bf 0.067 & 0.054 & 0.051 & 0.055 & 0.045 & 0.047 \\
D27 & 0.064 & 0.061 & 0.057 & 0.061 & 0.065 & 0.061 & 0.065 & \bf 0.083 & 0.069 & 0.064 & 0.071 & 0.069 \\
D29 & 0.052 & 0.048 & 0.045 & 0.047 & 0.052 & 0.054 & 0.048 & 0.055 & \bf 0.064 & 0.052 & 0.053 & 0.048 \\ 
D34 & 0.061 & 0.054 & 0.042 & 0.066 & 0.057 & 0.054 & 0.058 & 0.057 & 0.060 & \bf 0.081 & 0.049 & 0.046 \\
\bottomrule
\end{tabular}}
\end{table}

For a test of the proposed noise extractor's baseline performance, we conduct basic camera identification by correlating the obtained noise signals against the corresponding matching camera fingerprint estimate. A key question is whether each camera fingerprint warrants its own set of learned parameters. Table~\ref{tab:vision} gives some insight by reporting the median correlation scores under $H_1$ for the VISION database, where we employed all ten trained networks (arranged along the columns) for each probe image. For comparison, the last two columns list the corresponding results obtained from the standard Wavelet denoiser (Wav), and an off-the-shelf variant of the DnCNN~\cite{Zhang:2017aa} that we trained to extract Gaussian noise with standard deviation $\sigma=3$ on the 400 images provided by its authors. Notably, the SPN-CNN approach yields a measurable boost over the fingerprint-agnostic techniques when a camera-specific variant of the network is employed. The median correlation under $H_1$ increases, on average, by a factor of 1.5 (at a median correlation under $H_0$ of about $10^{-4}$ compared to $10^{-5}$ for the Wavelet denoiser), while camera-foreign network configurations yield $H_1$ correlations that are largely on par with prior art. The corresponding results for the six cameras from the Dresden Image Database in Table~\ref{tab:dresden} give a slightly different picture, as the network training seems to have less influence. It is currently unclear to what extent this is an artifact of the much more homogenous dataset (where each camera was set up to capture the same scenes \cite{Gloe:2010ad} and all images were processed with the same Adobe software). Images in the VISION dataset are more heterogenous, both in terms of content and camera-specific processing pipelines.\footnote{For instance, both device D06 and D15 are iPhone 6 cameras, but operate under different iOS versions.} More tailored datasets and experiments will be necessary to disentangle the impact of training data and sensor specifics. For the time being, we recommend training a camera-specific network for each fingerprint.

\begin{table}
\caption{Median correlation under $H_1$ ($400\times 400$ patches, Dresden).}%
\label{tab:dresden}%
\scriptsize\centering%
\vspace*{-4pt}\begin{tabular}{@{}c@{\ \ \ }*{7}{c@{\ }}@{\ }c@{}}
\toprule
& \multicolumn{6}{c}{SPN-CNN trained for camera} \\
\cmidrule(r){2-7}
\tikz[overlay]\node[rotate=90,xshift=.2cm,yshift=2pt]{camera};
\tikz[overlay]\node[rotate=0,xshift=.5cm,yshift=.55cm]{extractor};
\tikz[overlay]\draw[] (-.2,0.65) -- (.3,-0.1);
& D200-0 & D200-1 & D70-0  & D70-1  & D70s-0 & D70s-1 & Wav & \cite{Zhang:2017aa} \\
\cmidrule(r){2-7}
\cmidrule{8-9}
D200-0 & \bf 0.071 & 0.071 & 0.067 & 0.066 & 0.069 & 0.068 & 0.048 & 0.048 \\
D200-1 & 0.087 & \bf 0.089 & 0.084 & 0.083 & 0.086 & 0.084 & 0.059 & 0.060 \\
D70-0~  & 0.083 & 0.084 & \bf 0.088 & 0.087 & 0.087 & 0.086 & 0.053 & 0.052 \\
D70-1~  & 0.081 & 0.082 & 0.086 & \bf 0.086 & 0.086 & 0.084 & 0.050 & 0.050 \\
D70s-0 & 0.094 & 0.095 & 0.098 & 0.096 & \bf 0.098 & 0.096 & 0.059 & 0.059 \\
D70s-1 & 0.066 & 0.066 & 0.067 & 0.066 & 0.068 & \bf 0.068 & 0.040 & 0.040 \\
\bottomrule
\end{tabular}
\end{table}

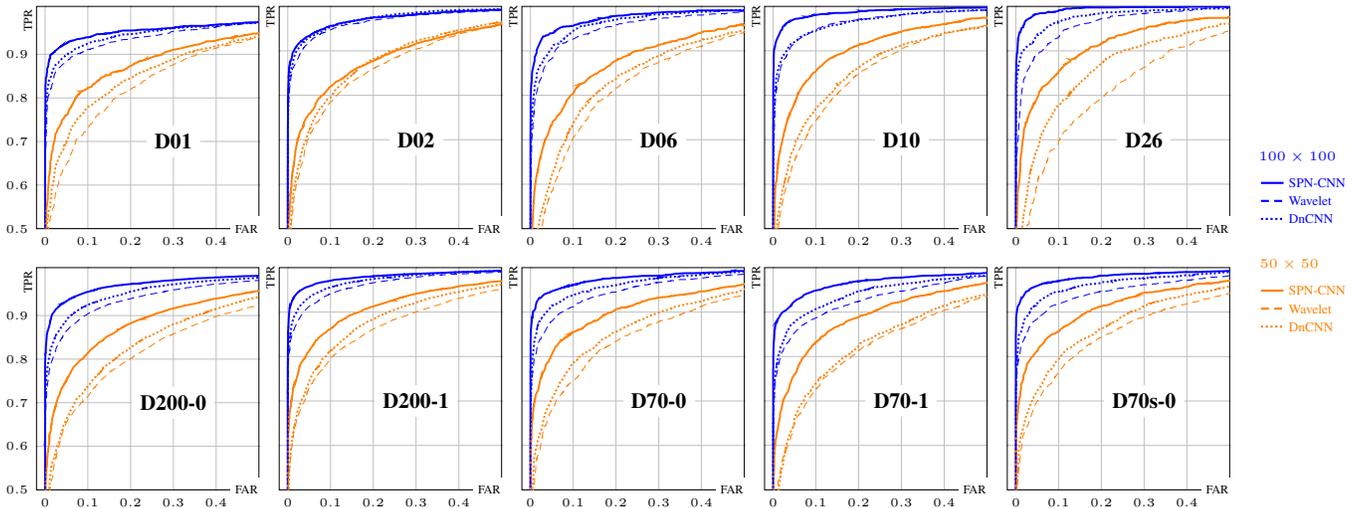
\begin{figure*}
\parbox{.9\textwidth}{\includegraphics[width=\linewidth]{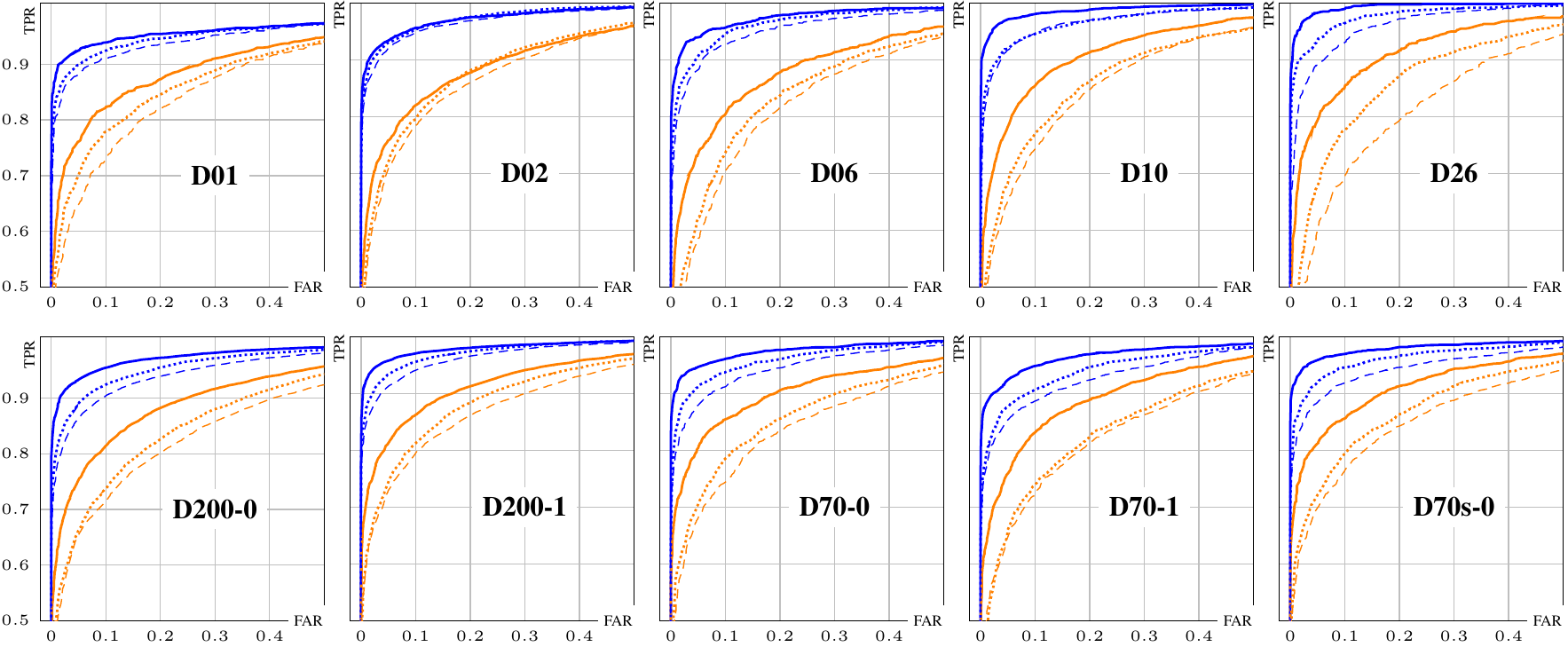}}\hfill%
\parbox{.08\textwidth}{\tiny%
\textcolor{blue}{$100\times 100$}\\[2ex]
\tikz\draw[blue,thick] (0,0) -- (.3,0) node[right]{SPN-CNN};\\
\tikz\draw[blue,thick,densely dashed] (0,0) -- (.3,0) node[right]{Wavelet};\\
\tikz\draw[blue,thick,densely dotted] (0,0) -- (.3,0) node[right]{DnCNN};\\[4ex]
\textcolor{orange}{$50\times 50$}\\[2ex]
\tikz\draw[orange,thick] (0,0) -- (.3,0) node[right]{SPN-CNN};\\
\tikz\draw[orange,thick,densely dashed] (0,0) -- (.3,0) node[right]{Wavelet};\\
\tikz\draw[orange,thick,densely dotted] (0,0) -- (.3,0) node[right]{DnCNN};\\[2ex]
}
\caption{Camera identification baseline ROC performance for selected cameras from the VISION Database (top) and the Dresden Image Database (bottom) based on the correlation between Wavelet-based  camera fingerprint MLE and noise signals extracted from probe patches of size $100\times 100$ (blue) and $50\times 50$ (orange) with SPN-CNN, and off-the-shelf Wavelet and DnCNN \cite{Zhang:2017aa} denoising, respectively.}
\label{fig:roc}
\end{figure*}

Adopting this premise, Figure~\ref{fig:roc} gives a more comprehensive picture of camera identification performance by reporting ROC curves for a select set of ten cameras from both datasets. We focus on smaller patches of size $100\times 100$ and $50\times 50$ here (center-cropped from the bigger probes) to showcase the advantage of the targeted data-driven extractors over conventional methods. Not surprisingly, smaller patch sizes incur a drop in performance, but the proposed approach achieves notable improvements across all tested cameras, including those not depicted here.

\subsection{Application: Manipulation Localization}
\label{sec:localization}

\begin{figure}[b]
\includegraphics{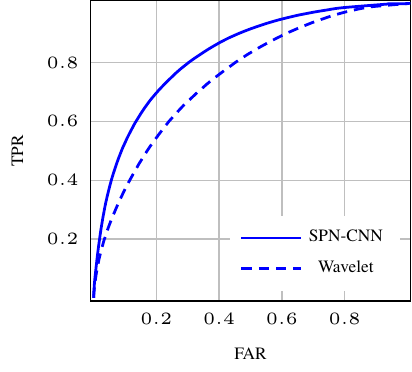}\hfill%
\includegraphics{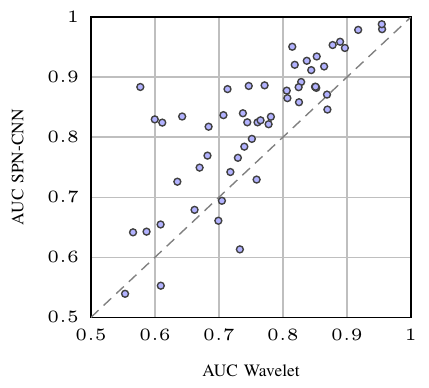}
\vspace*{-2ex}\caption{Image manipulation localization pixel-level ROC curves from 55 manipulated images (left) and per-image AUC scores (right) from correlating noise signals extracted with SPN-CNN and Wavelet denoising in sliding windows of size $64\times$64 with the Wavelet-based camera fingerprint MLE.}
\label{fig:localization}
\end{figure}

The promising performance on camera identification from small image patches suggests improved image manipulation localization. We test this conjecture by computing from the aforementioned 55 manipulated images $64\times 64$ sliding window correlations with the Nikon D7000 fingerprint estimate. Following prior art \cite{Chen:2008aa}, we also train a simple linear regression correlation predictor for both the SPN-CNN and the Wavelet noise extractors. A completely disjoint set of 20,000 patches from 20 images was used for this purpose. The pixel-level ROC curves (aggregated over all probe images) in the left panel of Figure~\ref{fig:localization} are obtained from thresholding the difference between measured and predicted correlation, $\Delta_i = \rho_i - \hat{\rho}_i$, i.\,e., a strongly negative difference is indicative of the camera fingerprint being absent in the respective local neighborhood. In line with the baseline results in Figure~\ref{fig:roc}, the graphs suggest that the data-driven SPN-CNN extractor is beneficial over the standard Wavelet approach by a large margin. The overall area under the curve (AUC) increases from 0.75 to 0.82. As the results generally vary greatly from image to image, we include a scatter plot of per-image pixel-level AUC scores in the right panel of Figure~\ref{fig:localization}. Observe that SPN-CNN performs better for all but seven probe images, which qualitatively seemed to exhibit content characteristics that were under-represented during training. In general, it is worth pointing out that the localization performance can be expected to increase further with one of the more sophisticated recent random field approaches \cite{Chierchia:2014ab,Chakraborty:2017aa,Korus:2017aa}.

\subsection{Application: Video Source Attribution}
\label{sec:video}

Video source attribution is generally more challenging than camera identification from still images. The final experiment of this initial exploration thus puts emphasis on the impact of lossy video compression and reduced resolution. We do not consider video stabilization here to keep the discussion focussed, but we anticipate that any performance boost in the absence of stabilization would ultimately also help techniques designed to reverse the effects of inter-frame geometric misalignment~\cite{Mandelli:2019aa}. To address the specific characteristics of video data, we train a separate network on frames extracted from the five training videos, while the target fingerprint was computed from the available still images \cite{Iuliani:2019aa}. As a result, synchronizing the imagery to the camera fingerprint requires special attention. The available videos have a resolution of $1280\times 720$, whereas the full sensor resolution of the still images is $2560\times 1920$ pixels. We use the guidance in \cite{Shullani:2017aa} to determine the geometric mapping from image to video space, but found that the cropping parameters had to be adjusted to reflect differences between landscape and portrait mode. Once camera fingerprint estimate and video frames are aligned, the training proceeds as before.

\begin{figure}
\includegraphics{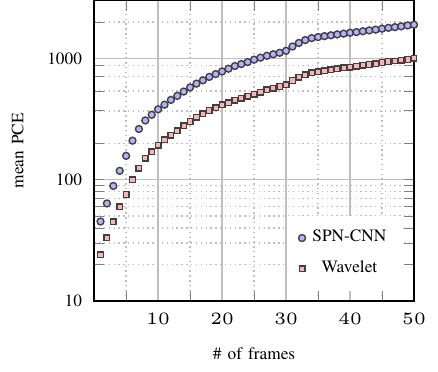}\hfill%
\includegraphics{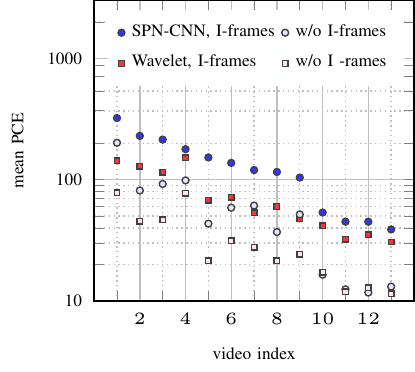}
\vspace*{-2ex}\caption{Video source attribution from the first $N$ probe video frames (left) and from individual frames (right). Average PCE scores from 13 unstabilized videos (left) and per video (right).}
\label{fig:video}
\end{figure}

The results in Figure~\ref{fig:video} offer two complementary perspectives. The left panel reflects the common procedure to aggregate noise estimates from multiple frames into a video fingerprint. The graphs depict the mean PCE score (averaged over all test videos) against the number of frames considered (starting from the first frame). The data-driven SPN-CNN approach again provides a measurable boost over the Wavelet denoiser, indicating the possibility of more reliable video source attribution. For a closer look at video-specific results, the right panel of Figure \ref{fig:video} reports average frame-level PCE scores per video, distinguishing between I-frames and all remaining frames.\footnote{The training did not make this distinction.} \mbox{I-frames} are expectedly more beneficial for source attribution, and noise signals extracted with the SPN-CNN resemble the camera fingerprint more closely. The average frame-level PCE over all I-frames increases from 75 to 135. 

Interestingly, there is a sharp drop in SPN-CNN performance for the last four videos compared to a more graceful decline of the Wavelet approach. A closer inspection revealed that these are all ``outdoor'' videos (each two from the ``move'' and ``panrot'' categories), with content that was not well represented during training. Along with a number of the earlier observations, we thus find it appropriate to close with a note of caution. While a data-driven perspective clearly holds great promise for boosting sensor noise forensics under challenging conditions, these advances may not come cheap in practice. Proper and careful training is crucial, and a single ``catch-all'' framework will likely require substantial further experimentation.

\section{Concluding Remarks}
\label{sec:conclusion}

We have demonstrated that digital camera identification from PRNU sensor pattern noise can benefit greatly from training a convolutional neural network to extract noise signals from probe images that resemble the expected camera fingerprint more closely than noise residuals obtained with standard fingerprint-agnostic denoising procedures. The discussed network features a clean end-to-end design that draws from the recent DnCNN residual learning approach \cite{Zhang:2017aa}. In its current instantiation, it achieves its most favorable results with a dedicated set of parameters for each candidate camera fingerprint, but future research questions abound. More research is needed to understand to what extent the apparent dependence on sensor-specifics is related to differences in the content of the data presented to the network. Along those lines, it is worth pointing out that we have not explicitly controlled for effects such as JPEG compression quality etc., and that practical applications may warrant a rigorous in-depth analysis of the estimators under $H_0$. Looking forward, it seems also reasonable to assume that the current approach may only be a stepping stone towards a fully data-driven camera identification framework. Fingerprint estimation and noise signal extraction may be learned jointly, likely leading to further performance boosts. This may address some of the recent concerns regard the viability of the fundamental imaging model that is also the foundation of this present work \cite{Masciopinto:2018aa}. Finally, questions pertaining to fingerprint-copy and removal attacks in the realm of counter-forensics \cite{Bohme:2012aa} will also have to be reconsidered when moving to data-driven approaches \cite{Bonettini:2018aa}.


\section*{Acknowledgment}

This work was supported by AFRL and DARPA under Contract No. FA8750-16-C-0166. Any findings and conclusions or recommendations expressed  in this material are solely the responsibility of the authors and does not necessarily represent the official views of AFRL, DARPA, or the U.S. Government.

\bibliographystyle{IEEEtran}
\bibliography{references}

\end{document}